\documentclass[preprint,12pt]{elsarticle_mine}

\usepackage{amssymb}
\usepackage{microtype}
\usepackage{booktabs}

\usepackage{xcolor}

\journal{Computer Vision and Image Understanding}

\begin{document}

\begin{frontmatter}



% \title{Exocentric To Egocentric Transfer For Action Recognition: A Short Survey}

\title{Egocentric and Exocentric Methods: A Short Survey}


\author[inst1]{Anirudh Thatipelli}
\author[inst2]{Shao-Yuan Lo}
\author[inst3]{Amit K. Roy-Chowdhury}

\affiliation[inst1]{organization={University of Central Florida}}

\affiliation[inst2]{organization={Honda Research Institute USA}}

\affiliation[inst3]{organization={University of California, Riverside}}

\begin{abstract}
Egocentric vision captures the scene from the point of view of the camera wearer, while exocentric vision captures the overall scene context. Jointly modeling ego and exo views is crucial to developing next-generation AI agents. The community has regained interest in the field of egocentric vision. While the third-person view and first-person have been thoroughly investigated, very few works aim to study both synchronously. Exocentric videos contain many relevant signals that are transferrable to egocentric videos. This paper provides a timely overview of works combining egocentric and exocentric visions, a very new but promising research topic. We describe in detail the datasets and present a survey of the key applications of ego-exo joint learning, where we identify the most recent advances.  With the presentation of the current status of the progress, we believe this short but timely survey will be valuable to the broad video-understanding community, particularly when multi-view modeling is critical.
\end{abstract}


\begin{keyword}
Egocentric \sep exocentric \sep ego-exo learning \sep action recognition.
\end{keyword}

\end{frontmatter}


\section{Introduction} \label{sec:intro}
Humans perceive the world from multiple viewpoints. We watch Do-it-yourself videos to learn new skills. A bicycle repair video alternates between the ego (first-person) and exo (third-person) viewpoints. An ego (close-up) view of the bicycle captures vital hand-object interactions and an exo (third-person) view captures the overall context in the environment. We can relate the object from the third-person to the first-person perspective. Being able to map skills to one's own body has been a well-studied problem in cognitive science \cite{flavell, newcombe, rizzolatti2004mirror}. Capturing video from both the \textbf{ego} and \textbf{exo} views is a vital frontier for AI to understand human activities. Widespread applications exist in augmented reality \cite{8951899, jimaging10010003} and robotics \cite{kang2023video, mu2016effect, RECCHIUTO201643}.

\begin{figure}[!t]
    \centering
    \includegraphics[width=1\linewidth]{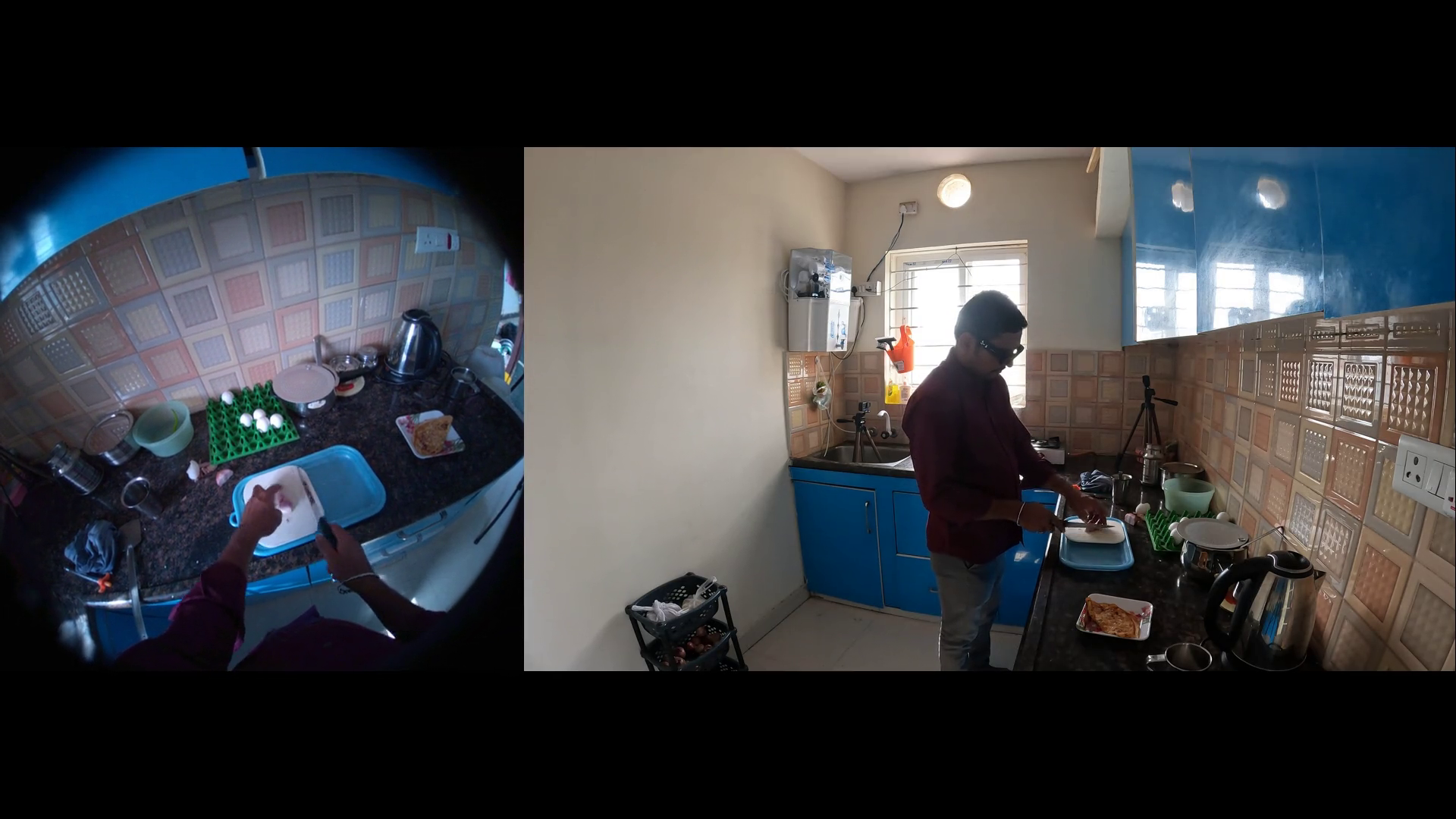}
    \caption{
        Hand-object interactions in the third-person view (right) are useful for identifying the action from the first-person viewpoint (left).
    }
    \label{fig:figure_main_ego_exo}
\end{figure}

Despite the importance of multi-view learning, most efforts into video understanding have focused on only one view, third-person (exocentric) viewpoint \cite{ucf, caba2015activitynet, carreira2022shortnotekinetics700human,agrawal2020person, agrawal2021deepsct, lal2023temp3d} or first-person (egocentric) viewpoints \cite{grauman2022ego4d} separately. While existing algorithms perform considerably well on third-person settings \cite{kin700classificationresults}, a significant gap exists in the egocentric settings \cite{grauman2024ego, damen2018scaling, epic-kitchens-100}.

\begin{figure}[!t]
    \centering
    \includegraphics[width=1\linewidth]
    {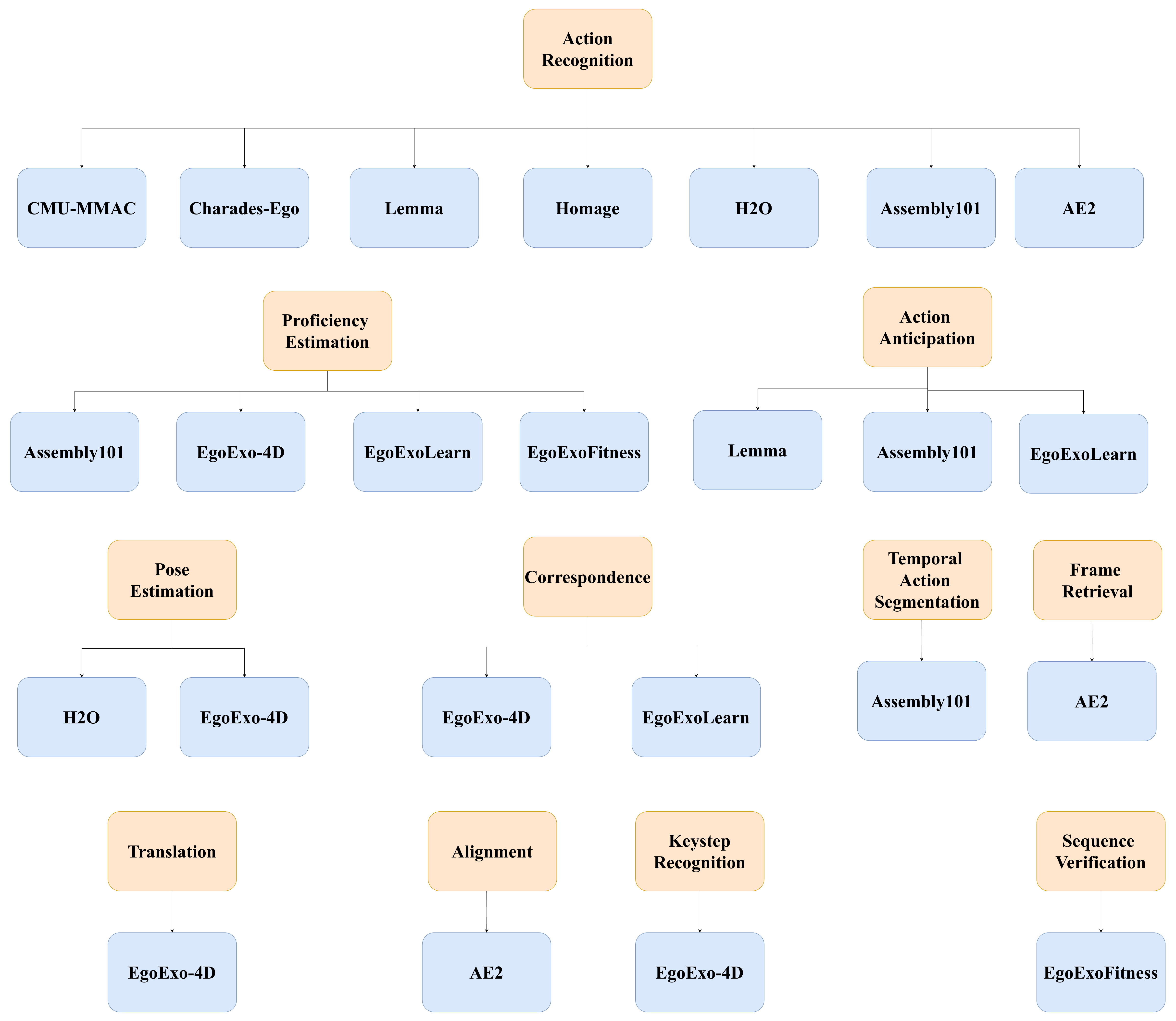}
    \caption{Ego-Exo Datasets and corresponding tasks.
     This figure illustrates the different Ego-Exo datasets in the literature and compares them to the associated benchmarks. Newly released Ego-Exo4D \cite{grauman2024ego}, EgoExoLearn \cite{huang2024egoexolearn}, EgoExoFitness \cite{li2024egoexofitnessegocentricexocentricfullbody} constitute a large suite of novel tasks to further research in this arena. 
    % \rohit{You can make the boxes bigger and text bigger. Use the full page}
    }
    \label{fig:figure_main}
\end{figure}

An exocentric view contains many relevant cues for recognizing the egocentric view. For example, in Figure~\ref{fig:figure_main_ego_exo}, the hand-object interaction of ``cutting'' in the third-person view can be useful to recognize in the first-person view. This short survey provides a timely high-level overview of various egocentric-exocentric learning tasks. Figure~\ref{fig:figure_main} provides an overview of these tasks.

The rest of this paper is organized as follows. In Section~\ref{sec:datasets}, we introduce the datasets that contain paired ego-exo views. In Section~\ref{sec:related_work}, we discuss existing approaches that have recently been presented for various applications. Finally, Section~\ref{sec:conclusion} concludes the paper with a summary and provides several promising future research directions.

\section{Datasets} \label{sec:datasets}
Several datasets containing paired ego-exo views have been proposed in the literature \cite{Kwon_2021_ICCV, cmu-kitchens, homage, sener2022assembly101, charades-ego}. ``Mixed'' ego-exo views have been covered by \cite{howto100m, tang2020comprehensive, youcook2, crosstask, yagi2023finebio, ben2021ikea, corona2021meva, Kuehne_2014_CVPR, xue2023learning}. Zhang \textit{et al.} \cite{zhang2022egobody} captures egocentric interactions in a 3D viewpoint. Xue \textit{et al.}'s AE2 dataset \cite{xue2023learning} is sampled from existing ego and exo datasets. These datasets have several shortcomings: lack of magnitude, weak synchronization, and poor diversity. The recently released large-scale datasets, such as Ego-Exo4D  \cite{grauman2024ego}, EgoExoLearn  \cite{huang2024egoexolearn}, and EgoExo-Fitness \cite{li2024egoexofitnessegocentricexocentricfullbody}, attempt to bridge this gap. Please refer to Table~\ref{tab:dataset_summary} for an overview.

Beyond the standard parameters outlined in Table~\ref{tab:dataset_summary}, these datasets differ in several key aspects, such as data modalities (e.g., RGB, depth, IMUs - Inertial Measurement Units, stereo), annotation richness (e.g., fine-grained vs. coarse), and the quality of synchronization between the egocentric and exocentic views. While some datasets are specifically designed for action recognition, others prioritize pose estimation or sequence verification. Additionally, datasets vary in their environmental diversity, ranging from controlled lab-settings to real-world outdoor scenarios.

\begin{table}[!t]
\centering
\caption{Comparison of existing datasets across various parameters, arranged in chronological order. }
\label{tab:dataset_summary}
\resizebox{\linewidth}{!}{
    \begin{tabular}{ccccccccc}
    \toprule
    Dataset & Publication Year & Hours & \# Action Clips & \# Scenarios & \# Subjects & \# Verb Classes & \# Noun Classes & \#Action Classes\\
    \toprule
    CMU-MMAC \cite{cmu-kitchens} & 2009 & - & 5 & 1 & 43 & - & - & 8\\
    Charades-Ego \cite{charades-ego} & 2018 & 68.8 & 68536 & 15 & 112 & 33 & 36 & 157\\
    Lemma \cite{jia2020lemma} & 2020 & 10.1 & 324 & 15 & 8 & 24 & 64 & 641\\
    Homage \cite{homage} & 2021 & 25.4 & 453 & 70 & 27 & 29 & 86 & -\\
    H2o \cite{Kwon_2021_ICCV} & 2021 & 5 & 500 & 36 & 4 & 11 & 8 & 36\\
    Assembly101 \cite{sener2022assembly101} & 2022 & 513 & 4321 & 15 & 53 & 24 & 90 & 1380\\
    AE2 \cite{xue2023learning} & 2023 & - & 322 & 6 & - & - & - & 4\\
    Ego-Exo4D \cite{grauman2024ego} & 2024 & 1286 & 5035 & 43 & 740 & 1481 & 2924 & 689\\
    EgoExoLearn \cite{huang2024egoexolearn} & 2024 & 120 & 747 & 8 & - & 95 & 254 & 39\\
    EgoExo-Fitness \cite{li2024egoexofitnessegocentricexocentricfullbody} & 2024 & 31 & 1248 & 1 & 49 & - & - & 76\\
    \midrule
    \end{tabular}
}
\end{table}

\textbf{CMU-MMAC dataset} \cite{cmu-kitchens} is one of the earliest datasets that captures ego and exo video. It is composed of 43 participants cooking 5 recipes in the kitchen setting. Multiple modalities like audio, video, accelerations, and motion capture are present in this dataset.

\textbf{Charades-Ego dataset} \cite{charades-ego} was one of the former large-scale joint multi-view dataset efforts containing 68.8 hours of first and third-person video. 112 actors hired by Amazon Mechanical Turk recorded 34 hours of scripted scenarios.

\textbf{Lemma dataset} \cite{jia2020lemma} comprises multi-view and multi-agent daily-life activities. 3D skeletons and RGB-D are collected to give a broad perspective.

\textbf{Homage dataset} \cite{homage} is a synchronized multi-view dataset comprising 30 hours of ego-exo video from 27 participants performing household activities in the same environment. It is well annotated with both the hierarchical and atomic action labels.

\textbf{H2O dataset} \cite{Kwon_2021_ICCV} focuses on 3D egocentric object-level manipulations. It comprises 3D hand poses, 6D object poses, camera poses, object meshes, and scene-point clouds. 4 different participants perform 36 unique actions in three unique environments.

\textbf{Assembly101 dataset} \cite{sener2022assembly101} features non-scripted multi-step activities. 101 toy vehicles are manipulated in 4321 video sequences for a total of 513 hours. It constitutes 1380 fine-grained and 202 coarse-grained action classes. \textbf{AssemblyHands} \cite{ohkawa2023assemblyhands} is a subset of Assembly101 to study the challenging problem of 3D hand pose estimation and action classification.

\textbf{AE2 dataset} \cite{xue2023learning} is one of the premier attempts to learn a view-invariant self-supervised embedding from unpaired ego and exo videos. To this end, they create a new benchmark, sampled from five public datasets \cite{epic-kitchens-100, cmu-kitchens, kuehne2011hmdb, Kwon_2021_ICCV, zhang2013actemes}, and a self-collected tennis dataset. It is composed of 322 clips.

\textbf{Ego-Exo4D dataset} \cite{grauman2024ego} is the largest multi-view dataset including the egocentric view and the corresponding exocentric information. Moreover, it also offers multiple natural language descriptions including expert commentary, narrate-and-act descriptions, and atomic action descriptions. It is rich in modalities like audio, IMUs, video, depth, gaze, stereo, 3D environments, thermal IR, GPS, motion capture, 6DOF, barometer and magnetometer readings. 740 subjects shot 123 scenes across different cities. It releases new challenging benchmarks like keystep recognition, efficient action detection, and proficiency estimation.

\textbf{EgoExoLearn dataset} \cite{huang2024egoexolearn} is another concurrent large-scale ego-exo synchronized dataset. It contains 120 hours of demonstration activities recorded in the lab and daily-life settings. It is richly annotated with fine-grained captions. Unlike previous datasets, it releases benchmarks on cross-view action anticipation and proficiency estimation.

\textbf{EgoExo-Fitness dataset} \cite{li2024egoexofitnessegocentricexocentricfullbody} was also concurrently released along with the previous two ego-exo datasets. While the previous datasets extensively explored daily-life activities, EgoExo-Fitness focuses on exercise-related activities. It comes with a new set of benchmarks for cross-view sequence verification.

\section{Existing Approaches} \label{sec:related_work}
\begin{figure}[!t]
    \centering
    \includegraphics[width=1\linewidth]{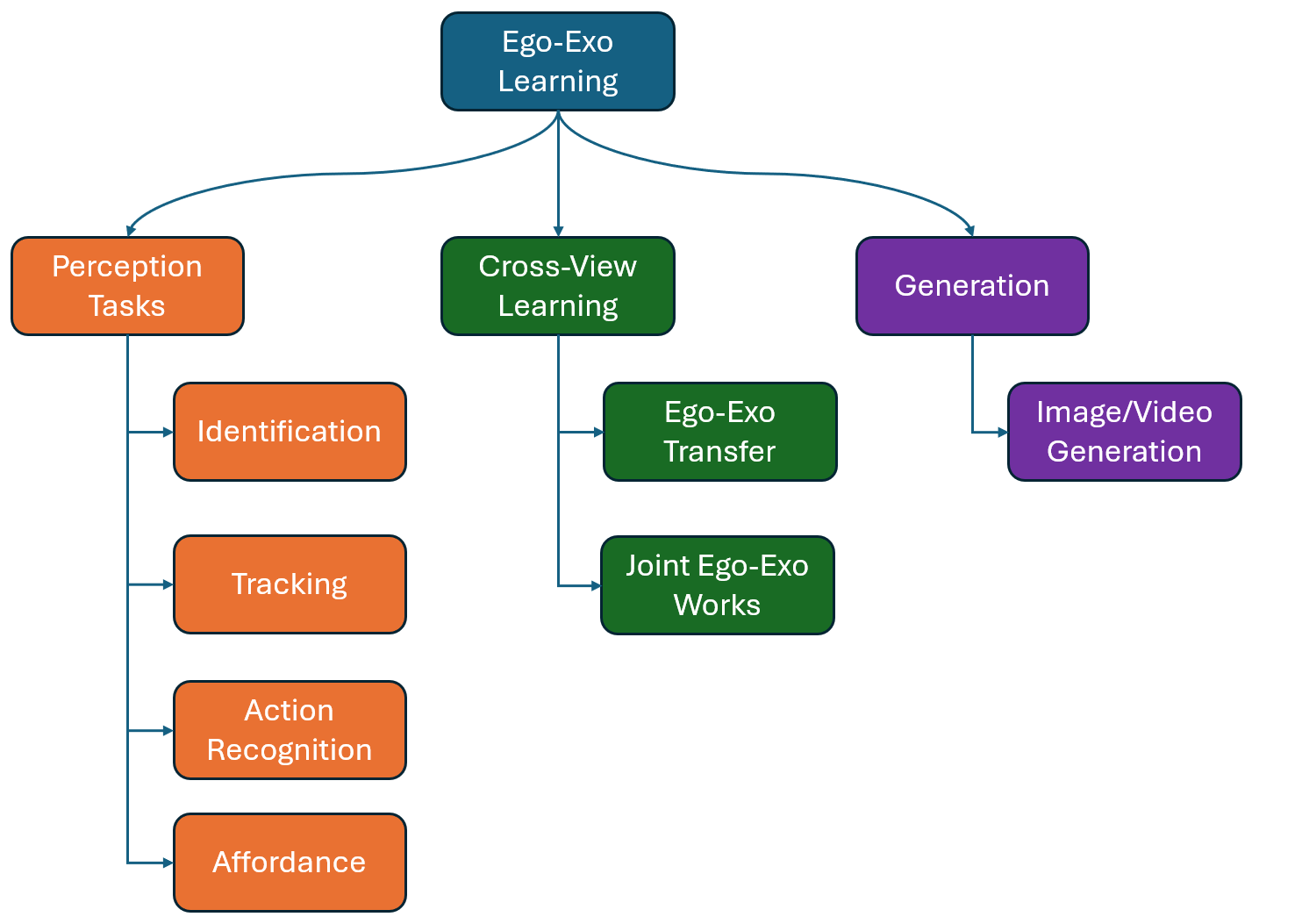}
    \caption{Categorization of the joint egocentric and exocentric tasks. These tasks can be grouped into three major groups: \textbf{Perception Tasks}, \textbf{Cross-View Learning} and \textbf{Generation}. Perception Tasks include identification, tracking, action recognition, and affordance analysis, while Cross-View Learning encompasses ego-exo transfer and joint ego-exo works.}
    \label{fig:figure_overview}
\end{figure}

Egocentric vision focuses on camera-wearer-centric cues while exocentric vision focuses on a broader perspective of the subject in the context of the entire scene. Leveraging the complementary signals from both viewpoints will enable us to learn human skills effectively.

Some early work has investigated the task of jointly relating egocentric and exocentric vision \cite{alahi2008object, soran2015action}.
In this section, we discuss important tasks jointly modeling vision from the first-person and third-person perspectives. In the following subsections, we will introduce existing approaches for various joint egocentric and exocentric applications, as illustrated in Figure~\ref{fig:figure_overview}.

\subsection{Identification}

\begin{figure}[!t]
    \centering
    \includegraphics[width=1\linewidth]{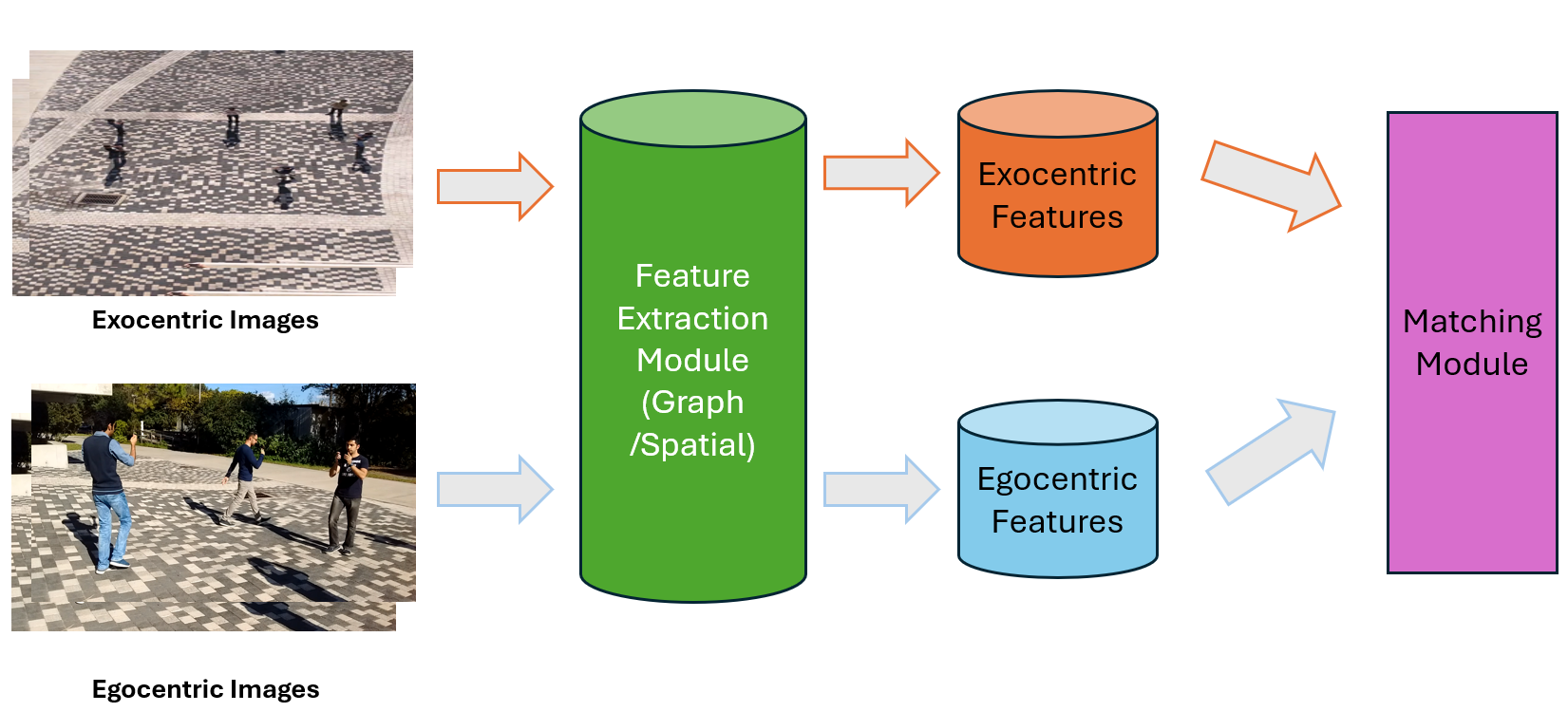}
    \caption{A modular framework for identification: A feature extraction module (graph/spatial) captures representations from the image inputs and fed into a matching module for identification. Images adapted from \cite{ardeshir2016ego2top}.}
    % \vspace{mm}
    \label{fig:egoexo_ident}
\end{figure}

It is the task of matching camera wearers in an egocentric video to an exocentric video. The lack of visibility in the egocentric video makes this task challenging. Being able to match a participant in both views is an important preliminary task for joint ego-exo learning. It is a well-researched problem. Ardeshir \textit{et al.} \cite{ardeshir2016ego2top, ardeshir2018egocentric} is one of the first works that proposes a graph-matching technique to solve this problem. Ardeshir \textit{et al.} \cite{ardeshir2018egocentric, Ardeshir_2018_ECCV} further propose a joint approach to tackle temporal alignment and person re-identification. Figure \ref{fig:egoexo_ident} presents a unified framework behind these models.
% Figure~\ref{fig:ego2top} shows some examples of this dataset.  
Similarly, Han \textit{et al.} \cite{han2019multiple} propose a different matching function based on spatial distributions. Fan \textit{et al.} \cite{fan2017identifying} learns a joint-embedding space. The model proposed by Ardeshir \textit{et al.} \cite{ardeshir2018integrating} extended to focus on temporal alignment. In Han \textit{et al.} \cite{han2020complementarycoview}, a conditional random field is proposed to identify the subjects from different viewpoints. Xu \textit{et al.} \cite{xu2018joint} perform simultaneous matching and segmentation of the subject across both views.

% \begin{figure}[!t]
%     \centering
%     \includegraphics[width=1\linewidth]{Ego2Top_2.png}
%     \caption{Egocentric views and the corresponding exocentric view (top view in this example) images taken from Ego2Top \cite{ardeshir2016ego2top}}. 
%     % \vspace{mm}
%     \label{fig:ego2top}
% \end{figure}

While identification focuses solely on matching the camera wearer, re-identification aims to learn the associations between the different subjects in the egocentric and exocentric views. Work by Ardeshir \textit{et al.} \cite{ardeshir2016egoreid} is one of the earliest approaches to exploring the task of re-identification between the different views. To enable further research in multi-view video-based re-identification, Basaran \textit{et al.} \cite{basaran2018egoreid} release a novel multimodal dataset, consisting of around $176,000$ detections. Han \textit{et al.} \cite{han2019multiple} utilizes spatial information such as the view-angle of the camera to perform the association. Han \textit{et al.}  \cite{han2022connecting} attempts to solve a challenging version of the problem by assuming limited appearance matching and different viewing angles in the ego and exo image.  Han \textit{et al.} \cite{han2023relating} considers another challenging variant of this matching problem having minimal overlap of the field-of-view.

While current methods assume shared features between ego and exo views, real world cases involve significant distortions and occlusions. Future models should explore occlusion-aware architecture and multi-modal fusion (e.g., depth, IMU sensors) to improve robustness. Further, exploring temporal feature consistency across frames could enhance re-identification accuracy.

\begin{figure}[!t]
    \centering
    \includegraphics[width=1\linewidth]{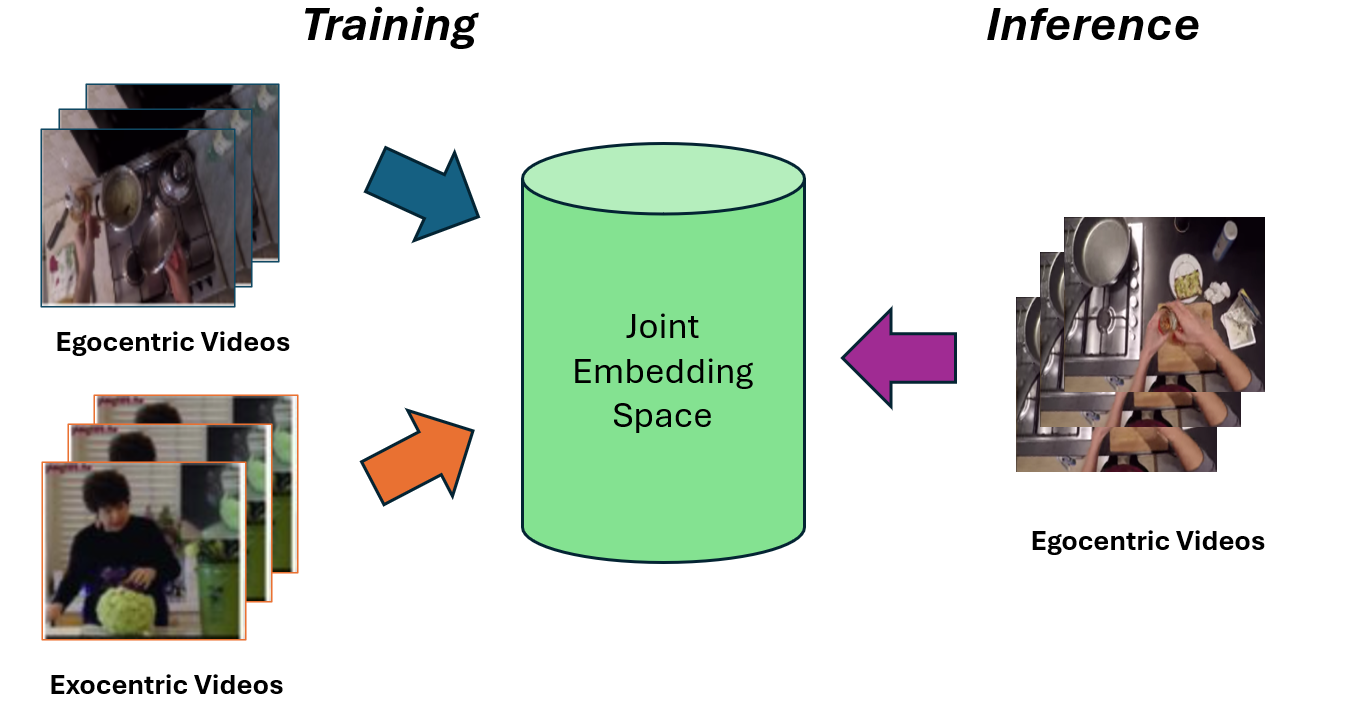}
    \caption{A unified framework for joint egocentric-exocentric action recognition. A joint multi-view representation space is learned from the egocentric and exocentric views. At inference, the model utilizes this learned representation to make egocentric video predictions, leveraging knowledge from exocentric views for improved recognition. Images taken from \cite{damen2018scaling} and \cite{kay2017kinetics}.}
    % \vspace{mm}
    \label{fig:action_recognition}
\end{figure}

\subsection{Action Recognition}

Action Recognition is the task of identifying or assigning a category or multiple categories to the action performed by the subject in the video. The release of GoPro wearable cameras significantly increased the production of first-person videos. However, limited works have combined ego and exo views for identifying actions. The earliest attempt to recognize human activity across first and multiple third-person cameras was done in \cite{soran2015action}. It presents a learnable weighted importance classification approach. Truong \textit{et al.} \cite{truong2023cross} learns a geometric constraint to transfer knowledge between the multiple views. Rocha \textit{et al.} \cite{rocha2023cross} learns an invariant space, based on skeleton pose information.  Huang \textit{et al.} \cite{huang2021holographic} extends to a multi-domain scenario, learning a holographic feature space based on both view-invariant and view-specific features. In Peng \textit{et al.} \cite{peng2021vvs}, virtual features from a first-person perspective are synthesized and combined to perform action recognition. Different from other works, Ramirez \textit{et al.} \cite{ramirez2017added} incorporates gaze information into the robot's internal representation for improved imitation of human behavior. 

Existing methods rely only on appearance-based features, which are sensitive to occlusions and viewpoint variations. Developing robust multi-modal feature representations (e.g., depth, IMU sensors, etc. can enhance action recognition. Overall, a unified framework of ego-exo action recognition is illustrated in Figure \ref{fig:action_recognition}.

\subsection{Tracking}

\begin{figure}[!t]
    \centering
    \includegraphics[width=1\linewidth]{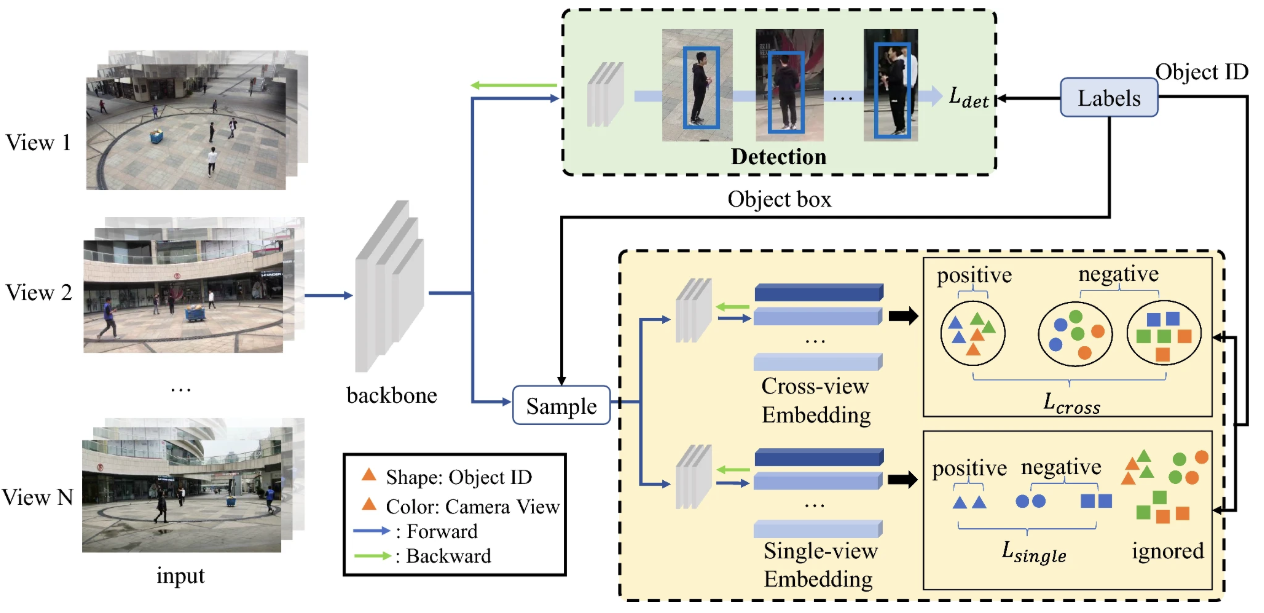}
    \caption{Illustration of a tracking framework from \cite{hao2024divotrack}, showcasing a unified approach for object detection, cross-view identity matching, and single-view tracking.}
    % \vspace{mm}
    \label{fig:tracking}
\end{figure}

Tracking is a fundamental Computer Vision problem, that involves estimating the global trajectories and match subjects across the video. Yang \textit{et al.} \cite{yang2019visual} is one of the earliest works that jointly identifies and tracks the subjects across the first and third-person views. A deep neural network (DNN), robust to action and motion changes is used to generate the 3D trajectory. Han \textit{et al.} \cite{han2020complementary} learns a spatio-temporal correspondence between the images of different viewpoints. In a follow-up work, Han \textit{et al.} \cite{han2020complementarycoview} treats the task as a joint optimization problem. Han \textit{et al.} \cite{han2024benchmarking} extends the optimization approach for relating a single third-person view with multiple first-person view images. Recent work by DivoTrack \cite{hao2024divotrack} presents a new baseline. CrossMOT for multi-view object tracking, illustrated in Figure \ref{fig:tracking}. Multi-view tracking has also gained importance in other areas like robotics \cite{lin2023joint}.

Most methods rely on manual annotations, making scalability challenging. Future research should explore self-supervised object tracking objectives to mitigate the need for labelled data. Secondly, developing occlusion-aware tracking models is crucial for robustness. Thirdly, current models are computationally expensive and unsuitable for deployment. Finally, multi-modal fusion by incorporating depth sensors and IMUs can enhance the generalization to diverse environments.

% \begin{figure}[!t]
%     \centering
%     \includegraphics[width=1\linewidth]{CVMHT.jpg}
%     \caption{Example ego-exo frames taken from the CVMHT dataset released in \cite{han2020complementary}. Images are taken from \cite{han2020complementary}.} 
%     % \vspace{mm}
%     \label{fig:tracking}
% \end{figure}

% \begin{figure}[!t]
%     \centering
%     \includegraphics[width=1\linewidth]{elfeki_dataset_frames.png}
%     \caption{Pairs from simultaneously recorded Ego-Top and Ego-Side dataset. Image taken from \cite{elfeki2018third}.} 
%     % \vspace{mm}
%     \label{fig:generation}
% \end{figure}

\begin{figure}[!t]
    \centering
    \includegraphics[width=1\linewidth]{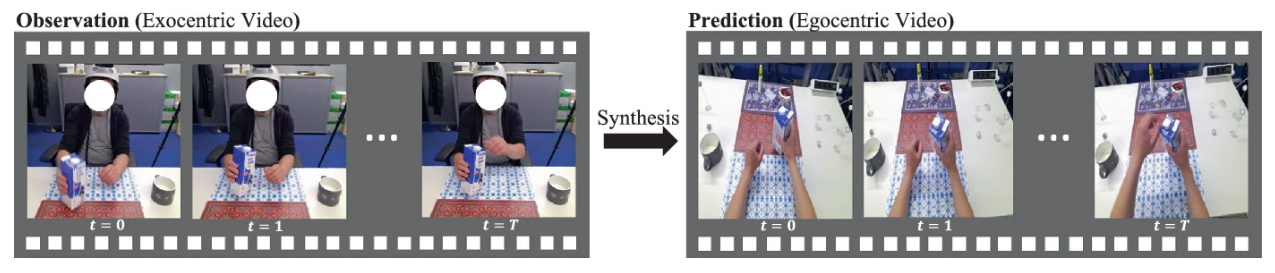}
    \caption{Illustration of the task that synthesizes egocentric video frames from exocentric video frames, as demonstrated in \cite{luo2024put}.}
    % \vspace{mm}
    \label{fig:generation}
\end{figure}

\subsection{Generation}

In generation, we aim to synthesize an egocentric image, conditioned on an exocentric image and vice versa. Elfeki \textit{et al.} \cite{elfeki2018third} was the first landmark dataset for exo-ego synthesis and retrieval. A conditional GAN \cite{mirza2014conditional} is used to synthesize first-person images.  Liu \textit{et al.} \cite{liu2020exocentric} also utilizes a variation of a GAN. Similarly, Tang \textit{et al.} \cite{tang2019multi, tang2022multi} utilize semantic information to generate images in different views. Liu \textit{et al.} \cite{liu2022parallel} utilize a shared network between the ego-exo frames to aid generation. Liu \textit{et al.} \cite{liu2021cross} synthesize egocentric videos by combining the semantic map with GANs. Recent work by Luo \textit{et al.} \cite{luo2024put} presents a diffusion-based approach \cite{ho2020denoising} for synthesizing egocentric videos from exocentric inputs, as illustrated in Figure \ref{fig:generation}. Different from all the other works, Luo \textit{et al.} \cite{luo2024intention} uses action description and egocentric frames to synthesize a video from the third-person perspective. The new Ego-Exo4D dataset \cite{grauman2024ego} constitutes a benchmark for synthesis. 

Significant progress has also been made in a related problem of aerial view to ground view synthesis \cite{regmi2018cross, toker2021coming, swerdlow2024street}.

Despite these advancements, several challenges remain unresolved. Bridging the viewpoint shift between egocentric and exocentric frames continues to be a major obstacle, as current models struggle to generalize across unseen perspectives. Integrating multi-modal signals, such as audio or IMU data, could enhance the realism of generated sequences and improve the scalability and efficiency for real-world deployment. Addressing these challenges will be crucial for advancing the field and enabling more robust, adaptable, and scalable solutions for egocentric-exocentric synthesis.

\subsection{Affordance}

\begin{figure}[!t]
    \centering
    \includegraphics[width=1\linewidth]{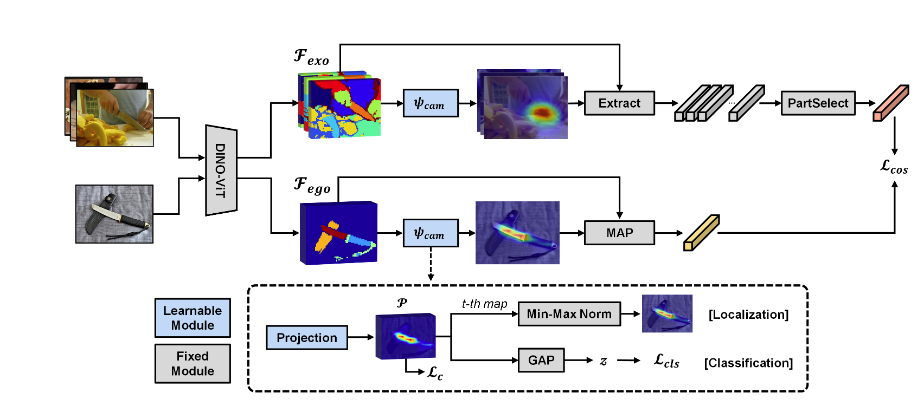}
    \caption{LOCATE framework \cite{li2023locate} utilizes DINO-ViT \cite{caron2021emerging} features for precise localization of human-object interactions.}
    % \vspace{mm}
    \label{fig:affordance}
\end{figure}

Much attention has been drawn to affordance \cite{gibson1977theory, jianjiasurvey, ardon2020affordances}. The objective is to understand the different possible actions that can be performed with an object. Luo \textit{et al.} \cite{luo2022learning, luo2023grounded} extracts affordance-level features from exocentric human-object interactions and transfers it to the egocentric view. Li \textit{et al.} \cite{li2023locate} extend this work using a weakly-supervised technique, as illustrated in Figure~\ref{fig:affordance}. Chen \textit{et al.} \cite{chen2023affordance} extends affordance learning from videos using an attention-based network. Xu \textit{et al.} \cite{xu2024weakly} also uses a weakly-supervised technique leveraging cross-view knowledge. Recent work by Zhang \textit{et al.} \cite{zhang2024self} integrates a self-explainable module to aid affordance learning. Yang \textit{et al.} \cite{yang2024learning} presents a joint coarse and fine-grained feature extraction technique. Different from other techniques, Rai \textit{et al.} \cite{rai2024strategies} leverage knowledge of Vision-Language Models (VLMs) as an auxiliary mask for the task of grounding. 
% Please see Figure~\ref{fig:affordance} for examples of images and the corresponding affordances.

A primary challenge, unaddressed in affordance learning methods is occlusion of critical objects which can obscure important affordance cues. Additionally, variability is introduced due to the diversity of human-object interactions. Finally, integrating multi-modal sensor data, such as depth data and IMUs can provide additional context to affordances. Addressing these challenges will advance the field of affordance learning, particularly in applications involving egocentric and exocentric perspectives.

\subsection{Exo-Ego Transfer}
\label{sec:exo_ego_transfer}

\begin{figure}[!t]
    \centering
    \includegraphics[width=1\linewidth]{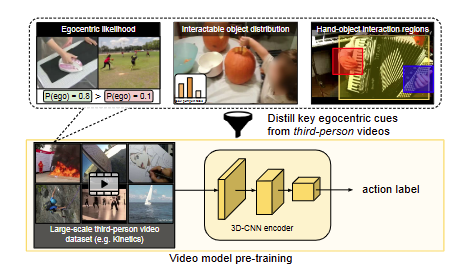}
    \caption{Ego-Exo approach for transferring egocentric cues from exocentric videos \cite{li2021ego}.}
    % \vspace{mm}
    \label{fig:ego_exo_transfer}
\end{figure}

A vast amount of knowledge in the form of motion cues is embedded in exocentric videos that can be transferred to the egocentric domain. Transferring knowledge from exocentric to egocentric views present several key challenges, primarily due to significant viewpoint mismatches, occlusions, and differences in motion representation. Ardeshir \textit{et al.} \cite{ardeshir2016egotransfer} is a premier work that learns mappings between the ego-exo views. In Ardeshir \textit{et al.} \cite{ardeshir2018exocentric}, the authors propose a two-stream view-specific architecture to adapt from exo to ego view. Ho \textit{et al.} \cite{ho2018summarizing}  utilizes a semi-supervised domain adaptation technique to adapt exocentric visual cues to egocentric videos. Xu \textit{et al.} \cite{xu2023pov} uses a prompt-masking technique for transferring information for egocentric hand-object interaction. These domain adaptation techniques require expensive large scale-datasets. Unlike earlier methods, Li \textit{et al.} \cite{li2021ego} introduce an enhanced training strategy that effictively extracts signals from exocentric videos for application in the egocentric domain, as shown in Figure \ref{fig:ego_exo_transfer}. Ohkawa \textit{et al.} \cite{ohkawa2023exo2egodvc} aids further adaptation by performing view-invariant pretraining and finetuning. Different from previous techniques, Quattrocchi \textit{et al.} \cite{quattrocchi2023synchronization} proposes an adaptation technique for temporal action segmentation. 

In Nishimura \textit{et al.} \cite{nishimura2023view}, the geometric transformation is used to tackle a novel problem of view-verification (bird's eye-view trajectory estimation) is computed from the egocentric movement. However, this approach may encounter challenges in real-world scenarios where the camera placement is inconsistent. Qian \textit{et al.} \cite{qian2024bird}
is an extension to a more challenging problem of bird's eye view estimation in the absence of proper calibration. 

\subsection{Joint Ego-Exo Works}

\begin{figure}[!t]
    \centering
    \includegraphics[width=1\linewidth]{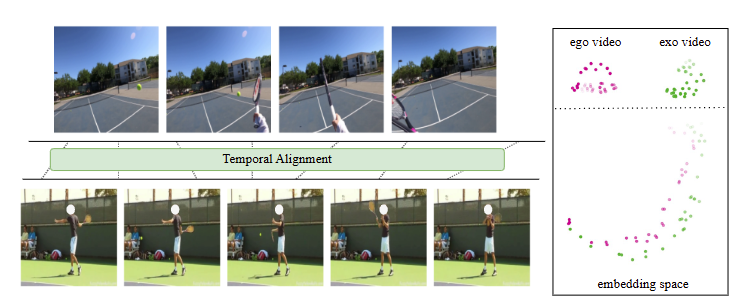}
    \caption{Temporally aligned view-invariant representation learned from the ego-exo videos, as proposed by \cite{xue2023learning}.}
    % \vspace{mm}
    \label{fig:joint_ego_exo}
\end{figure}

This section outlines works that aim to learn a joint ego-exo representation. Sigurdsson \textit{et al.} \cite{sigurdsson2018actor} makes the first attempt to jointly relate first-person and third-person viewpoints. Yu \textit{et al.} \cite{yu2019see, yu2020first} leverages a joint attention mechanism to extract a shared representation between the views. These approaches are reliant on learning an attention model, making them computationally intensive. In Wang \textit{et al.} \cite{wang2023learning}, a sentence-bert language model \cite{reimers2019sentence} is utilized to semantically align the unpaired exocentric and egocentric videos. Xue \textit{et al.} \cite{xue2023learning} became the first work to propose a self-supervised learning approach to learn a view-invariant representation. Figure \ref{fig:joint_ego_exo} illustrates the approach. Zhao \textit{et al.} \cite{zhao2024fusing} solves a novel task of identifying and segmenting the egocentric camera wearer in a third-person view. 

3D egocentric pose estimation has also benefited from a joint ego-exo learning framework \cite{dhamanaskar2023enhancing, liu2023egofish3d}. A novel thermal image-based 3D hand-pose dataset has been released in ThermoHands \cite{ding2024thermohands}. Lu \textit{et al.} \cite{lu2021multi} covers a scene-graph generation technique based on a self-attention mechanism between the ego and exo views. The authors of  Wen \textit{et al.} \cite{wen2021seeing} present a solution combining third-person and first-person images to predict the subject's location from the third-person viewpoint. Jia \textit{et al.} \cite{jia2024audio} extracts exocentric and egocentric conversational signals to generate a scene graph. Xu \textit{et al.} \cite{xu2024retrieval} shows an improvement in egocentric captioning by retrieving semantically relevant third-person videos \cite{asai2023retrieval}. Future research should explore more robust multi-view fusion techniques, better temporal modeling, and efficient, scalable approaches to address these challenges and unlock new possibilities for joint ego-exo applications.

\subsection{Miscellaneous Applications}

Jointly relating exocentric and egocentric vision has applications in Robotics and Virtual Reality. Kennedy \textit{et al.} \cite{kennedy2007spatial} illustrate the importance of combining egocentric and exocentric information for mapping. Multi-view visual feedback to the robots of the swarm can improve performance \cite{recchiuto2016visual}. This has been corroborated in robotics manipulation as well \cite{jangir2022look}. Supervision from third-person videos has been well-adapted to egocentric vision in robotics \cite{sermanet2018time, bahl2022human, spisak2024diffusing}. A combination of hand and third-person perspective has been used in \cite{hsu2022vision}. Young \textit{et al.} \cite{young2022effects} demonstrates superior performance on the aerial telemanipulation task using egocentric-exocentric views.  Abdullash \textit{et al.} \cite{abdullah2024ego} synthesize third-person view from first-person view for enhanced teleoperation. Video captioning has also benefited from joint ego-exo information \cite{kang2023video}.

Combining ego and exo views has been thoroughly researched in virtual reality \cite{dede2009introduction, milgram1999taxonomy, dunleavy2014augmented}. Multiple works \cite{grasset2005evaluation, rasmussen2019scenecam, yu2015hand} demonstrate the possibility of using a mixed viewpoint space for collaboration. Soares \textit{et al.} \cite{soares2018ego} proposes a novel cooperative virtual environment with fixed freedom of movement per user. Peschel \textit{et al.} \cite{peschel2012towards} illustrates the use-case of a joint ego-exo system for unmanned aerial systems. Automatic rendevous and docking (ARD) also benefit from a joint view system \cite{larson2023examination}. Duncan \textit{et al.} \cite{duncan2023fusing}'s work proposes a camera system to reconstruct embodied experiences in real-time.

In recent years, VLMs have shown their effectiveness on various vision tasks, such as action anticipation~\cite{mittal2024can,zhao2024antgpt}, anomaly detection~\cite{xu2025towards,yang2024anomalyruler}, and affective understanding~\cite{guo2024stimuvar,xie2024emovit}. Existing VLMs \cite{radford2021learning, jia2021scaling} trained on large amounts of third-person perspective contain many signals that can be transferable for egocentric tasks. Previous works like \cite{li2021ego, ardeshir2016egotransfer} utilize simpler architectures to learn egocentric representations from third-person data. VLMs would be more capable of learning stronger representations for more complicated joint ego-and-exo tasks.

\section{Conclusion and Future Directions} \label{sec:conclusion}
In this survey, we cover the various ego and exo datasets and compare them across various attributes. The release of large-scale datasets \cite{grauman2024ego, huang2024egoexolearn, li2024egoexofitnessegocentricexocentricfullbody} and the associated challenges mark an exciting period in the field. Furthermore, we survey various challenges covered in this field: identification, action recognition, tracking, generation, affordance, ego-exo transfer, joint ego-exo tasks, and miscellaneous applications.

Given the evolving research landscape, we propose several promising directions for future exploration in joint egocentric and exocentric applications. The details are shown below. \\

\begin{enumerate}
    \item \textbf{Fine-Grained Multi-View Correspondences}: A major challenge in joint ego-exo tasks is the accurate alignment across the first and third-person views. The viewpoint shift between ego and exo view introduces occlusions, geometric distortions, and ambiguities in the hand-object interactions. Current methods, reliant on feature-based similarity matching struggle when the viewpoints observe different object appearances and motion trajectories. Future research should explore 3D-aware representations to better model fine-grained correspondences between the egocentric and exocentric data.

    \item \textbf{Temporal dependencies across perspectives}: Existing works primarily focus on spatial relationships, while the temporal consistency acoss ego-exo views remains unexplored. Recent studies \cite{li2021ego} indicate the potential of temporal alignment techniques to enhance predictive modeling of long-term interactions and actions.Investigating self-supervised sequence alignment could enhance predictive long-term predictive modelling.

    \item \textbf{Robustness to noisy, misaligned, and adversarial inputs}: Practical deployment of ego-exo models in applications such as sports analytics, autonomous systems, and assistive AI requires robustness to real-world variations. Current models fail when faced with sensor noise, occlusions, dynamic lighting, and adversarial perturbations. Developing adaptive fusion mechanisms, self-supervised robustness objectives, or adversarial training tailored for multi-view learning could significantly enhance model generalization. Additionally, leveraging multi-modal cues (e.g., audio, depth, IMU data) could provide redundancy and improve stability in challenging environments.

    \item \textbf{Realtime and Efficient Processing}: Current models are computationally expensive, limiting real-world deployment in robotics, AR/VR and autonomous systems.

    \item \textbf{Dataset Collection and Annotation}: There is a need for even richer datasets that capture the diverse multi-view scenarios across challenging environments. Additionally, capturing synchronized ego-exo data at a higher frame rate could facilitate better temporal alignment.

    \item \textbf{Multi-modal fusion and sensor integration}: Incorporating multi-modal data such as sensors, IMUs, and LiDAR can enhance robustness to occlusions, varying lightning conditions, and motion blur.
    
\end{enumerate}

By addressing these challenges, future research can enable more generalizable, temporally aware, pratical, and robust ego-exo models, unlocking new possibilities for human activity recognition, immersive AR/VR, and robotics.

% Firstly, investigating fine-grained multi-view correspondences, such as hand-object interactions is crucial for enhancing scene understanding and action recognition. Developing methods that align the interactions across multiple viewpoints can enhance the generalization and robustness of these methods. Secondly, while existing methods emphasize  spatial relationships, integrating temporal dependencies remains and unsolved challenge. Capturing the evolution of actions across perspectives is essential for modeling long-term interactions and improving predictive capabilities. Lastly, ensuring model robustness against real-world challenges, including noisy data, misalignment, and adversarial perturbations, is critical for practical deployment.  Investigating these mechanisms will facilitate the reliable application of joint ego-exo models across various domains such as human activity analysis, assistive AI, and autonomous systems.}

 \bibliographystyle{elsarticle-num} 
 \bibliography{survey}





% \end{thebibliography}
\end{document}